\documentclass{article} %
\usepackage{iclr2022_conference,times}

\usepackage{amsmath,amsfonts,bm}

\def\eqref#1{equation~\ref{#1}}

\def\1{\bm{1}}

\DeclareMathAlphabet{\mathsfit}{\encodingdefault}{\sfdefault}{m}{sl}
\SetMathAlphabet{\mathsfit}{bold}{\encodingdefault}{\sfdefault}{bx}{n}

\usepackage{wrapfig}
\usepackage{hyperref}
\usepackage{graphicx}
\usepackage{bbm}
\usepackage{algorithm}
\usepackage[noend]{algpseudocode}
\usepackage{amssymb}
\usepackage{booktabs}
\usepackage{mathabx}
\DeclareMathOperator{\mean}{mean}

\newcommand{\rev}[1]{{#1}}

\title{Batch-Softmax Contrastive Loss\\ for Pairwise Sentence Scoring Tasks}

\author{Anton Chernyavskiy\\
HSE University\\
Moscow, Russia \\
\texttt{aschernyavskiy@gmail.com} \\
\And
Dmitry Ilvovsky\\
HSE University\\
Moscow, Russia \\
\texttt{dilvovsky@hse.ru}\hspace{82pt} \\
\AND
Pavel Kalinin\\
Yandex\\
Moscow, Russia \\
\texttt{kalinin.pavel.v@gmail.com} \\
\And
Preslav Nakov\\
Qatar Computing Research Institute, HBKU\\
Doha, Qatar \\
\texttt{pnakov@hbku.edu.qa}\\
\AND
}

\iclrfinalcopy %
\begin{document}

\maketitle

\begin{abstract}
The use of contrastive loss for representation learning has become prominent in computer vision, and it is now getting attention in Natural Language Processing (NLP).
Here, we explore the idea of using a batch-softmax contrastive loss when fine-tuning large-scale pre-trained transformer models to learn better task-specific sentence embeddings for pairwise sentence scoring tasks.
We introduce and study a number of variations in the calculation of the loss as well as in the overall training procedure; in particular, we find that data shuffling can be quite important.
Our experimental results show sizable improvements on a number of datasets and pairwise sentence scoring tasks including classification, ranking, and regression.
Finally, we offer detailed analysis and discussion, which should be useful for researchers aiming to explore the utility of contrastive loss in NLP.
\end{abstract}

\section{Introduction}
\label{sec:intro}

Recent years have seen a revolution in Natural Language Processing (NLP) thanks to the advances in machine learning. While a lot of attention has been paid to the architectures, especially for deep learning, there has been less focus on studying loss functions.
At the same time, loss functions based on similar or on the same ideas were reinvented multiple times under different names. This can cause difficulties when solving new problems or when designing new experiments based on previous results. To a greater extent, this applies to ``universal'' loss functions, which can be applied in different machine learning areas and tasks such as Computer Vision (CV), Recommendation Systems, and NLP. An example of such universal loss function is the batch-softmax contrastive (BSC) loss, which we will discuss below. Our contributions can be summarized as follows:
\begin{itemize}
    \item We study the use of a batch-softmax contrastive loss for fine-tuning large-scale transformers to learn better task-specific sentence embeddings for pairwise sentence scoring tasks.
    \item We introduce and study a number of variations in the calculation of the loss such as symmetrization, incorporating labeled negatives, aligning scores on the similarity matrix diagonal, normalizing over the batch axis, as well as in the overall training procedure, e.g.,~shuffling, trainable temperature, and sequential pre-training.
    \item We demonstrate sizable improvements for a number of pairwise sentence scoring tasks such as classification, ranking, and regression.
    \item We offer detailed analysis and discussion, which would be useful for future research.
\end{itemize}

\section{Related Work}
\label{sec:related}

The contrastive loss was proposed by \citet{1640964} as metric learning that contrasts Euclidean distances between embeddings of samples from one class and between samples from different classes. \citet{NIPS2005_a7f592ce} suggested the triplet loss, which is based on a similar idea, but uses triplets (\textit{anchor}, \textit{positive}, \textit{negative}), and aims for the difference between the distances for  (\textit{anchor}, \textit{positive}) and for (\textit{anchor}, \textit{negative}) to be larger than a margin.
$N$-pair loss was presented as a generalization of the contrastive and the triplet losses as a way to solve the problem of extensive construction of hard negative pairs and triplets \citep{NIPS2016_6b180037}.

To this end, a batch of $N$ pairs of examples from $N$ different classes is sampled, and the first element in each pair is considered to be an \textit{anchor}. Thus, for each \textit{anchor}, there are one positive and $N - 1$ negative pairs. The loss contrasts the distances simultaneously using the softmax function over dot-product similarities. The approach was used successfully in computer vision (CV) tasks. The same method of Multiple Negative Ranking for training Dot-Product Scoring Models was applied to ranking natural language responses to emails \citep{Henderson2017EfficientNL}, where the loss uses labeled pairs. A similar idea, called \emph{Negative Sharing}, was used to reduce the computational cost when training recommender systems \citep{10.1145/3097983.3098202}. \citet{Wu2018UnsupervisedFL} presented an approach with $N$-pairs like logic, as a Non-Parametric Softmax Classifier, replacing the weights in the softmax with embeddings of samples from such classes. It was also proposed to use L2 normalization and temperature. \citet{yang-etal-2018-learning} proposed to use Multiple Negative Ranking to train general sentence representations on data from Reddit and SNLI. 
\rev{\citet{logeswaran2018an} presented a Quick-Thoughts approach to learn sentence embeddings, which constructs batches of contiguous sets of sentences, and for each sentence, contrasts the next sentence in the text and all other candidates.}

A lot of subsequent work has focused on maximizing Mutual Information (MI). \citet{Oord2018RepresentationLW} presented a loss function based on Noise-Contrastive Estimation, called InfoNCE. It models the ``similarity'' function that estimates the MI between the target (future) and the context (present) signals, and maximizes the MI between temporally nearby signals. If this ``similarity'' function expresses the dot-product between embeddings, the InfoNCE loss is equivalent to the $N$-pair loss up to some constants. It was also shown that InfoNCE is equivalent to the Mutual Information Neural Estimator (MINE) up to a constant \citep{pmlr-v80-belghazi18a}, whose minimization maximizes a lower bound on MI. Deep InfoMax (DIM) \citep{hjelm2019learning} improves MINE, and can be modified to incorporate some autoregression as InfoNCE. However, \citet{DBLP:conf/iclr/TschannenDRGL20} pointed out that the effectiveness of loss functions such as DIM and InfoNCE might be primarily connected not to deep metric learning but rather to MI.

The idea gained a lot of popularity in Computer Vision with the advent of SimCLR (a Simple framework for Contrastive Learning of visual Representations), which introduced NT-Xent (normalized temperature-scaled cross-entropy loss) \citep{pmlr-v119-chen20j}. It uses self-supervised learning, where augmentations of the same image are considered as positive examples and augmentations of different images are used as negative examples. Thus, the task is as follows: for each example in a batch, find its paired positive augmentation. Here, the $N$-pairs loss is modified with a temperature parameter and with an L2 normalization of embeddings to the unit hypersphere. The loss was further extended for supervised learning as SupCon loss \citep{khosla2020supervised}, which aggregates all positive examples (from the same class) in the softmax numerator.

\rev{Subsequently, these losses were introduced to the field of Natural Language Processing (NLP). \citet{Gunel2020SupervisedCL} combined the SupCon loss with the cross-entropy loss and obtained state-of-the-art results for several downstream NLP tasks using RoBERTa. \citet{Giorgi2020DeCLUTRDC} and \citet{DBLP:journals/corr/abs-2005-12766} used NT-Xent to pre-train Transformers, considering spans sampled from the same document and sentences augmented with back-translation as positive examples. \citet{Luo2020CAPTCP} proposed to use NT-Xent in a self-supervised setting to learn noise-invariant sequence representations, where sentences augmented with masking were considered as positive examples.
Finally, \citet{gao2021simcse} introduced the SimCLR loss to NLP under the name SimCSE (Simple Contrastive Learning of Sentence Embeddings), where sentences processed by a neural network with dropout served as augmentations of the original sentences. Here, we explore various ways to use a similar loss function for pairwise sentence scoring tasks.}

While the above-described loss functions have different names, they are all based on similar ideas. Below, we will use the name \emph{Batch-Softmax Contrastive} (BSC) loss, which we believe reflects the main idea best. In our experiments below, we will use the ``modern'' variant of the loss: with temperature, normalization, and symmetrization components (described in more detail in Section \ref{sec:bsc}). These components were not used for NLP in combination before. We further introduce a number of novel and important modifications in the definition of the loss and in the training procedure, which make it more efficient, and we show that using the resulting loss yields better task-specific sentence embeddings for pairwise sentence scoring tasks.

\section{Method}
\label{sec:methods}

\subsection{Batch-Softmax Contrastive (BSC) Loss}
\label{sec:bsc}

Pointwise approaches for training models for pairwise sentence scoring tasks, such as mean squared error (MSE), are problematic as the loss does not take the relative order into account. For instance, for two pairs with correct target scores (0.4, 0.5), the loss function would equally penalize answers like (0.3, 0.6) and (0.5, 0.4). However, the first pair is better, as it keeps the correct ranking, while the second one does not. This is addressed in pairwise approaches, e.g.,~in triplet loss, where the model directly learns an ordering. Yet, there is a problem for constructing pairs or triplets in the training set, as it is hard to find non-trivial negatives examples.

\begin{figure}[!t]
\centering
\includegraphics[width=0.65\textwidth]{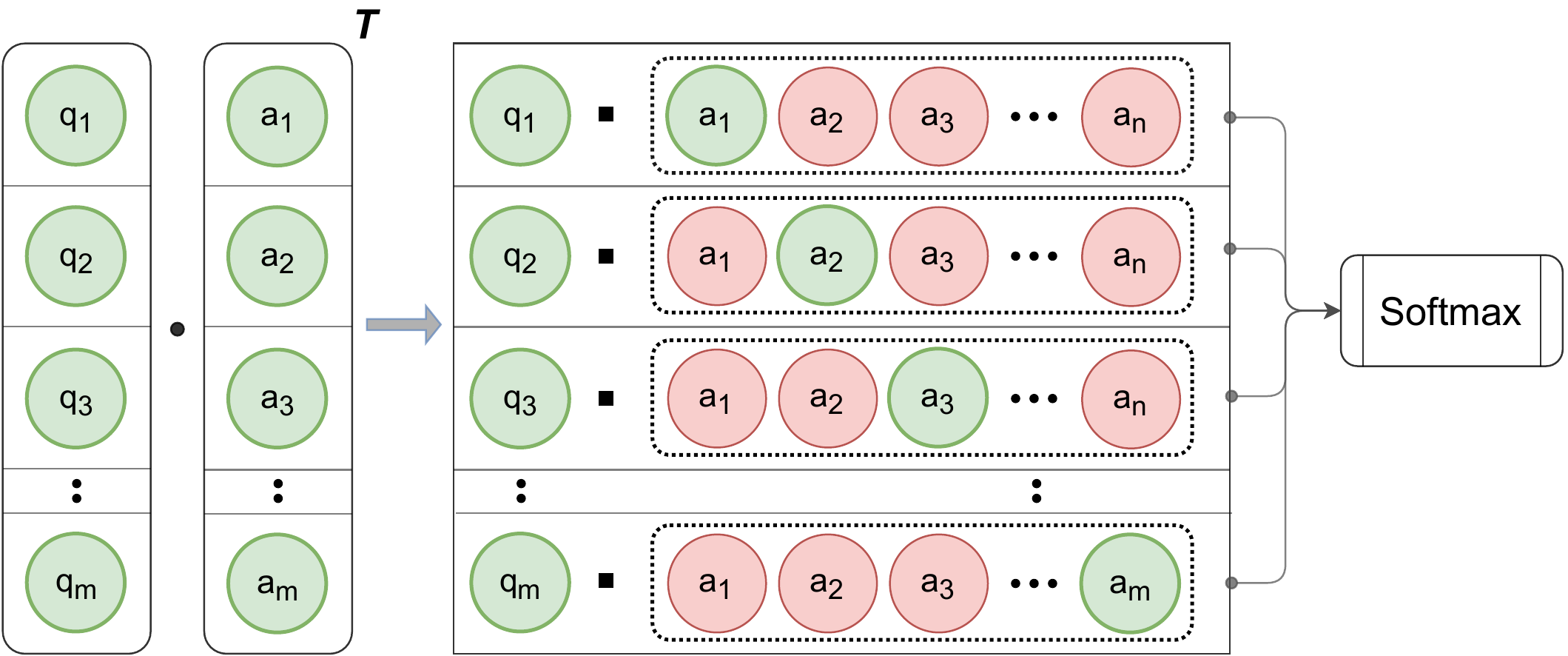}
\caption{For the set of positives pairs $(q_i, a_i)$, e.g.,~question--answer, for each $q_i$, the BSC loss contrasts the scores between $q_i$ and $a_i$ (positive examples) vs. between $q_i$ and $a_j$ for all $j \neq i$ (negative examples) using softmax. Here $\sqbullet$ denotes the dot-product.}
\label{fig_1}
\end{figure}

Unlike traditional pairwise loss functions, the BSC loss treats all other possible pairs of examples in the batch as ``negatives.'' That is, only positive pairs are needed for training. Consider a batch $X$ of pairs from a question-answering dataset. In general, let $Q_{m \times n}$ and $A_{m \times n}$ be the matrices of embeddings produced by a query model and an answer model. We define the loss function as follows: 
\begin{multline} \label{eq:1}
\mathcal{L}_{BSC}(X) = \mathcal{L}_0(X) + \mathcal{L}_1(X)  \\ =-\mean\left(\log \left (diag \left(\text{softmax}\left (\dfrac{QA^T}{\tau}\right)\right) \right)\right) -\mean\left(\log \left (diag\left(\text{softmax}\left (\dfrac{AQ^T}{\tau}\right)\right)\right)\right)
\end{multline}

Here, softmax is applied by rows (Figure~\ref{fig_1}), and $\tau$ is the temperature. Both components can be rewritten, e.g.,~$\mathcal{L}_0(X)$ can be written as follows:
\begin{equation} \label{eq:2}
\mathcal{L}_0(X) =
-\dfrac{1}{m\tau}\sum_{i=1}^{m}q_i^T a_i +\dfrac{1}{m}\sum_{i=1}^{m}\log\sum_{j=1}^{m}\exp\left (\dfrac{q_i^T a_j}{\tau}\right)
\end{equation}

Mathematically, this loss function is similar to the one presented in \citep{pmlr-v119-chen20j}. The difference is that we do not use augmentations, and we do not compare $q_i$ to $q_j$ (or $a_i$ to $a_j$) due to their different nature: we want to compare a question to an answer, not a question to a question or an answer to an answer. Thus, we apply the symmetrization in the formula. \rev{Note that, although a frequent short answer may fit multiple questions in a batch,  such pairs are considered as ``negative'' examples in the loss. However, the loss learns Mutual Information \citep{DBLP:conf/iclr/TschannenDRGL20}, that is $p(q_i, a_i)/(p(q_i)p(a_i))$, and thus it is robust to this false negatives problem.}

Early research has already shown the importance of properly configuring and using some BSC loss settings. For example, low temperatures are equivalent to optimizing for hard positives/negatives \citep{khosla2020supervised}, while L2 normalization of vectors to the unit hypersphere along with temperature effectively weighs different examples \citep{pmlr-v119-chen20j}. We further propose a number of important modifications that can have a major impact on the performance for a number of tasks.

\subsection{Batch Construction} 

In computer vision, it is common to use a batch size of 5,000, which in turn would naturally be very likely to contain some hard negative examples. In NLP, fine-tuning Transformer-based models with large batch sizes requires very large amounts of memory. Thus, much smaller batches are used in practice, and as a result, it becomes important to make sure these batches do contain some hard negative examples. We achieve this by fixing the content of the batches at each epoch of the training process. \rev{Note that this is much simpler than mining hard negatives, as we only need to increase the likelihood that there would be a hard negative example present in the batch, but we do not need to know which particular example in the batch would be hard. Inside the batch, this would be controlled by the temperature parameter.}

\textbf{Example-based shuffling}
The key idea of this method is to batch several groups, so that within each group all pairs are similar based on their first or based on their second elements. In this way, each positive pair would be accompanied by hard negatives from the same group and by simpler negatives from the remaining examples inside the batch (which come from other groups). We use the $k$-nearest neighbors for an input example to form a group for it, and Faiss \citep{8733051} to quickly find these nearest neighbors in the embedding space. Let the pairs be grouped by their first elements $q_i$. Algorithm~\ref{alg:1} summarizes the proposed method.

\begin{wrapfigure}{R}{0.52\textwidth}
\vspace{-20pt}
    \begin{minipage}{0.52\textwidth}
        \begin{algorithm}[H]
        \caption{Example-based shuffling}\label{alg:1}
        \begin{algorithmic}
            \State \textbf{Input:} sequence $D$, group size $s$
            \State initialize $R \leftarrow []$ \Comment{sequence to store the result}
            \State initialize $U \leftarrow \varnothing$ \Comment{set of used examples}
            \State randomly shuffle $D$
            \For{\texttt{e} in $D$}
                \If{\texttt{e} $\notin U$}
                \State find the $n$ nearest neighbors of \texttt{e} from $D$
                \State choose the top $s-1$ that are not in $U$
                \State add them and \texttt{e} to $R$ and also to $U$
                \EndIf
            \EndFor
            \State \textbf{return} reversed $R$
        \end{algorithmic}
        \end{algorithm}
    \end{minipage}
\end{wrapfigure}

Note that \rev{we use two stages in kNN to limit the range of possible candidates and thus to reduce the computational costs (both in terms of time and memory).} We first extract the top-$n$ neighbors (for some large $n$, e.g.,~500), and then we take the top-$k$ from them, so that no duplicates appear in the final sequence (for some small $k=7$). The time complexity of such a check is O(1). If all \rev{such} neighbors are already used, then only the considered example will be added to the resulting sequence. This case will often arise for the last examples, and thus batches will consist of simple 1-element groups. Therefore, we reverse the sequence to start with these simple batches, \rev{as in curriculum learning}.

\rev{By default, we assume that there should be one positive example for each question/answer (on the diagonal of the matrix)}, and thus identical neighbors could be optionally filtered. Still, if there are the same $q_i$ in the batch $X$, the loss definition (eq.~\ref{eq:1}) does not change. Indeed, let $P_q = \{i\,|\,q_i=q, (q_i, a_i) \in X\}$, then $\forall i,j \in P_q: (q_i, a_j)$ form a positive pair. According to \citet{khosla2020supervised}, \rev{for each $q$, all $\tilde q \in P_q$ should be placed in the softmax numerator and then averaging over all such $\tilde q$ should be performed outside the logarithm}. Thus, in $\mathcal{L}_0(X)$ (eq. \ref{eq:2}) only the first sum would change:
\begin{equation}
\sum_{i \in P_q}q_i^T a_i \rightsquigarrow \sum_{i \in P_q}\dfrac{1}{|P_q|}\sum_{j \in P_q} q_i^T a_j
= \sum_{j \in P_q}\dfrac{1}{|P_q|}\sum_{i \in P_q} q^T a_j = \sum_{j \in P_q}q_j^T a_j
\end{equation}

And in $\mathcal{L}_1(X)$:

\begin{equation}
\sum_{i \in P_q}q_i^T a_i \rightsquigarrow \sum_{i \in P_q}\dfrac{1}{|P_q|}\sum_{j \in P_q} q_j^T a_i
= \sum_{i \in P_q}\dfrac{1}{|P_q|}\sum_{j \in P_q} q^T a_i = \sum_{i \in P_q}q_i^T a_i
\end{equation}

To select the groups even better, we consider task-specific embeddings. To this end, we apply the current model to encode all pairs at each epoch.

\textbf{Fast shuffling} For extremely large datasets, example-based shuffling is time-consuming even with Faiss; thus, we propose several effective options to perform a less-thorough shuffling.
We choose some attribute by which we will group the examples, that is, we guarantee some closeness of the examples. Thus, the examples are close if they share the same words, the same cluster number or the same nearest neighbors. First, consider the case of words and grouping by the first elements of the pairs (the case of the second elements is the same). Algorithm~\ref{alg:2} presents the shuffling process.

\begin{algorithm}[H]
\caption{Shuffling by words}\label{alg:2}
\begin{algorithmic}
    \State \textbf{Input:} sequence $D$, group size $k$, shingle size $t$
    \For{\texttt{e} in $D$}
        \State \texttt{e.shingle} $\leftarrow$ random subset of $t$ words 
        \State \hspace{15pt}of \texttt{e} (\rev{ignoring stop-words})
    \EndFor
    \State sort $D$ by \texttt{e.shingle}
    \State initialize \texttt{gID} $\leftarrow$ random uint64 \Comment{group ID}
    \State initialize $s$ $\leftarrow$ 0 \Comment{current group size}
    \State initialize \texttt{prev} $\leftarrow$ first element of $D$
    \For{\texttt{e} in $D$}
        \If{\texttt{e.shingle} $\neq$ \texttt{prev.shingle}}
            \State \texttt{gID} $\leftarrow$ random uint64
        \EndIf
        \If{s $\geq$ k}
            \State \texttt{gID} $\leftarrow$ random uint64
            \State $s$ $\leftarrow$ 0
        \EndIf
        \State \texttt{e.gID} $\leftarrow$ \texttt{gID}
        \State $s \leftarrow s + 1$
        \State \texttt{prev} $\leftarrow$ \texttt{e}
    \EndFor
    \State sort $D$ by \texttt{e.gID}
    \State \textbf{return} $D$
\end{algorithmic}
\end{algorithm}

To produce a shuffle by clusters, we apply the same algorithm, where each sentence is replaced by its cluster number. Thus, each shingle has size $t = 1$. In order to make a shuffle by nearest neighbors, we create shingles by ``sentences,'' where the words are the positions of the top-$k$ nearest neighbors in the input sequence (for some small $k$). All of these approaches, as well as $k$-means clustering, can be effectively implemented using MapReduce and parallel computations. 

\subsection{Labeled Negatives} 

Usually, when the data size is small, hard negative examples may be hard to obtain even with data shuffling, e.g.,~when all examples are semantically distant. Nonetheless, if the dataset contains a labeled negative pair with some anchor, then its elements are semantically close by traditional rules of dataset construction. Thus, using such a pair inside the batch, where this anchor is present, will add the necessary hard negative example.

The only change that is added in the loss function is the masking of negative examples---we have no guarantees that the selected negative example is closer to the anchor than the rest of the examples inside the batch. Let $y_i$ be a binary label, where $y_i = 1$ if the $i$-th pair is positive. Then, we have
\begin{equation} \label{eq:3}
\mathcal{L}_0(X) = -\dfrac{1}{m\tau}\sum_{i=1}^{m}\mathbbm{1}[y_i=1]q_i^T a_i \\ +\dfrac{1}{m}\sum_{i=1}^{m}\mathbbm{1}[y_i=1]\log\sum_{j=1}^{m}\exp\left (\dfrac{q_i^T a_j}{\tau}\right)
\end{equation}

\subsection{Combo Loss}

Theoretically, it is beneficial to use several loss functions for training if they are calculated on the same batch (and thus do not require additional computations). That is, joint training of BSC and MSE losses combines the advantages of pointwise and of pairwise approaches, thus ensuring that for positives examples, the values on the diagonal of the dot-product matrix are not only greater than the rest, but are also close to 1 or to some target similarity. \rev{Note that here $L_{MSE}(X) = \frac{1}{m}\sum_i^m ((q_i^Ta_i) - y_i)^2$ for target positive similarities $y_i$, and thus we do not force all other similarities to zero.} At the same time, the BSC loss adds new examples (``negative'' pairs) to the training set.

In order to use the BSC loss when training a model in tasks with non-binary labels, we modify the indicator function in the equation \ref{eq:3}, as $\mathbbm{1}[y_i > t]$, where $t$ is a configurable binarization threshold. Then, we use their convex combination with the configurable hyperparameter $\mu \in (0,1)$:
\begin{equation}
    L(X) = \mu L_{BSC}(X) + (1 - \mu) L_{MSE}(X)
\end{equation}

\subsection{Normalization}

L2 normalization of %
matrices $A$ and $B$ means that $a_i^T b_j$ will be equivalent to cosine similarity.
The embeddings can also be normalized by the batch dimension (by coordinates), which can bring additional regularization. In our experiments, we confirm the importance of this, e.g.,~new representations can be calculated with L2 normalization by coordinates or in a min-max scale.

\section{Datasets}
\label{sec:datasets}

NLP tasks that compare pairs of sentences can be divided into regression (predicting a similarity score), classification (e.g.,~similar vs. dissimilar), and ranking (search for the best matches). They differ only by the quality assessment functions, and thus they all can benefit from the above losses.

Note that it is important to calculate sentence representations in ranking tasks, as when independently calculating the embeddings of the individual elements in the pairs, the inference time of the model becomes linear instead of quadratic. Therefore, we use Sentence-BERT (SBERT), which is trained as a Siamese BERT model, and offers a way to obtain state-of-the-art sentence embeddings, which have been proven useful for a number of tasks \citep{reimers-gurevych-2019-sentence, thakur2020augmented}. At inference time, we first use SBERT to obtain independently a representation for each sentence in the pair, and then we calculate the cosine similarities between these embeddings.

We use the following English datasets and tasks for the evaluation. Four ranking tasks (ranking answers to non-factoid questions, ranking questions by their similarity with respect to other questions, ranking comments by their similarity to a given question, ranking fact-checked claims by their relevance with respect to an input claim), two binary classification tasks (paraphrases identification, and duplicate question identification), and one regression task (semantic sentence similarity).

\textbf{Antique} The dataset contains 2,626 non-factoid questions with answer choices \citep{DBLP:journals/corr/abs-1905-08957}, asked by users on \emph{Yahoo! Answers}. There are a total of 34,011 question--answer pairs: 27,422 for training and 6,589 for validation. 
Each answer is annotated with a relevance score with respect to the question on a scale from 1 to 4,
and the task is to rank the answers by their relevance. To model relevance as a cosine similarity, we normalize the scores to the $[0, 1]$ interval. We use Mean Reciprocal Rank (MRR) as the main evaluation measure.

\textbf{CQA-A} This dataset was used in SemEval-2017 Task 3 on Community Question Answering subtask A \citep{nakov-etal-2017-semeval}.
The goal is to rank the first ten answers in a question thread on Qatar Living, so that \emph{good} answers are ranked higher than \emph{bad} ones. We used the \emph{clean} part of the dataset, which consists of 14,110 and 2,440 labeled question--comments pairs for training and development, respectively. The evaluation measure is Mean Average Precision (MAP). This dataset contains important metadata, e.g., the date and time of the comment, and sorting the comments by time yields a strong baseline; yet, we only use the text. To train the model with the triplet loss, we group the pairs by the first element (anchor).

\textbf{CQA-B} This dataset was developed for SemEval-2017 Task 3, subtask B \citep{nakov-etal-2017-semeval}, whose goal was to rank 10 potentially related questions by their similarity with respect to an input question. 
These questions are retrieved from the Qatar Living forum using Google and the input question as a query. We use the clean part of the dataset, which consists of 19,990 training and 5,500 development labeled question-question pairs. The main evaluation measure here is MAP. 
There is additional information, e.g.,~the rank of the retrieved question in the Google search results, which we do not use.

\textbf{PFCC-S} \citet{shaar-etal-2020-known} presented a dataset for detecting Previously Fact-Checked Claims on Snopes (PFCC-S), aimed at facilitating the solution of a fact-checking problem: given an input claim, it asks to rank claims that have been previously fact-checked, so that claims that can help verify the input claim or parts thereof are ranked as high as possible. The dataset has 800 positive input--verified claim pairs for training and 200 such positive pairs for testing, and they are to be matched against a database of 10,369 verified claims. The evaluation is performed in terms of a HasPositive@k metric, which checks whether there is a positive match among the first $k$ results in the ranked list. There are no negative examples in the dataset, but about eight million such pairs can be created. Thus, in order to train models using MSE or triplet loss, we sampled negatives according to the following scheme. First, we encoded all sentences from the database using SBERT, pretrained on STS and NLI. Then, we selected the first element in each positive pair as an anchor and we sorted all other examples by their similarity to this anchor. The assumption is that positive examples will be concentrated in the beginning, e.g.,~among the top-100. Thus, we selected negatives starting from 101 on, logarithmically: on positions $100 + 2^k, k \in \mathbbm{N}$. As a result, we obtain many hard negative examples and a small number of easy ones. Moreover, the balance between positive and negative examples is similar in order. Finally, we oversampled the positive pairs to correct the balance and to train with MSE or combined pairs into triplets by common anchors.

\textbf{Microsoft Research Paraphrase Corpus (MRPC)} \citet{10.3115/1220355.1220406} contains 5,800 pairs of sentences, extracted from online news sources. Each pair was labeled with a tag indicating whether the sentences are paraphrases (semantically equivalent). There are 3,668, 407, and 1,725 pairs in the training, development, and test subsets. As it is a binary classification task with class imbalance, it is evaluated in terms of F1.

\textbf{Quora Question Pairs (QQP)} Quora presented a dataset containing over 500,000 sentences with over 400,000 lines of potential duplicate questions.
Each line has a binary label indicating whether the line truly contains a duplicate pair. Due to the sampling method, which returns mostly positive pairs, the authors supplemented the dataset with negative pairs composed of ``related questions.'' As in \citep{thakur2020augmented}, we sample randomly 10,000 examples for training, and we use the F1 score as the main evaluation measure.

\textbf{Semantic Textual Similarity Benchmark (STSb)} The STS benchmark comprises a selection of the English datasets used in the STS tasks organized in the context of SemEval between 2012 and 2017 \citep{Cer_2017}. The benchmark comprises 8,628 sentence pairs. The pairs were annotated with similarity scores on a scale from 0 to 5 (5 indicating complete equivalence). There are a total of 5,749, 1,500 and 1,379 pairs in the training, in the development, and in the testing split, respectively. The main metric is Spearman's rank correlation. As in the Antique dataset, we normalize the scores to the $[0, 1]$ interval \rev{and then we binarized them based on a threshold of $0.6$}.

\section{Experimental Setup}
\label{sec:exp}

We used BERT-base uncased in all our experiments to be able to perform direct comparison for tasks such as MRPC, QQP and STS to previous work \citep{reimers-gurevych-2019-sentence, thakur2020augmented}. 

We set the number of warm-up steps to 10\% of the total steps, and we limited the input sequence length to 90 subtokens. We used a batch size of 30 in all tasks, except for Antique and QQP, where we used 50. Note that an order of magnitude larger batch sizes would probably yield better results, but they would also require much more memory. We experimented with learning rates from \{5e-6, 1e-5, 2e-5, 3e-5\}, and we selected (on dev) 3e-5 for CQA-B and 2e-5 for all other experiments. We used the AdamW optimizer with the bias correction for the CQA tasks, and without bias correction for the rest. We trained the model for five epochs for Antique, CQA-A and STSb, for six epochs for QQP, MRPC and PFCC-S, and for seven epochs for CQA-B, saving a checkpoint after each one, and we selected the best checkpoint on dev. As recommended in \citep{thakur2020augmented}, due to instability, we did seed optimization, running each approach five times and selecting the best result (on dev).

To train with the BSC loss, we used min-max normalization by coordinates with $\tau = 1.2$ for PFCC-S and QQP, standard L2 normalization with $\tau = 0.055$ for CQA-A, $\tau = 0.07$ for CQA-B, and $\tau = 0.1$ for all other tasks (to find the optimal $\tau$, we made it trainable for one run). We applied example-based shuffling to train with the BSC loss. We used a group size of four in MRPC, of five in CQA-B, and of eight in all other tasks. We iterated over $\mu$ values from the set $\{0.1, 0.5, 0.9\}$, and we chose $\mu = 0.1$ to train the \textit{combo} approach for CQA-A, MRPC, QQP and STSb tasks, and $\mu = 0.9$ for the other experiments. Unless otherwise stated in the results table, \textit{BSC} denotes using these settings.

We trained the triplet loss variant from \citep{reimers-gurevych-2019-sentence} with $\texttt{margin} = 0.6$ for PFCC-S, and $\texttt{margin} = 0.5$ for all other tasks. As we have no answers for the test set in MRPC, and no test sets in Antique and PFCC-S, we split the training set into 9:1 to tune the hyper-parameters. The time for training SBERT with the BSC loss (or \textit{combo} loss) was almost equal to the time for training with the standard MSE loss. We ran all experiments on a GeForce GTX 1080 GPU.

\section{Results}
\label{sec:results}

\begin{wraptable}{R}{0.52\textwidth}
\small
\centering
\vspace{-23pt}
\caption{Results for Antique.}
\begin{tabular}{l c c c}
\toprule
\textbf{Approach / Metric} & MRR & P@1 & nDCG@1  \\
\midrule
MSE & 0.781 & 0.660 & 0.769 \\
BSC & 0.804 & 0.680 & 0.754 \\
BSC - positives & 0.784 & 0.655 & 0.744 \\
BSC - random shuffle & 0.799 & 0.670 & 0.754 \\
Combo BSC + MSE & \bf 0.822 & \bf  0.710 & \bf 0.773 \\
\midrule
\citet{DBLP:journals/corr/abs-1905-08957} & 0.797 & 0.709 & 0.713 \\
\bottomrule
\end{tabular}
\label{antique}
\end{wraptable}

\textbf{Antique} The results are shown in Table \ref{antique}.  Our best approach of combo-training MSE and BSC losses outperforms all other variants and the approach proposed in \citep{DBLP:journals/corr/abs-1905-08957}, where specific negative sampling and a triplet loss were used.
Besides, the best BSC configuration achieves higher scores than MSE. We can see the importance of using predefined hand-crafted negative examples, which brings additional difficult cases and increases MRR by 0.02.

\begin{wraptable}{R}{0.55\textwidth}
\small
\centering
\vspace{-23pt}
\caption{Results for CQA-A and CQA-B.}
\begin{tabular}{l c c c c}
\toprule
\textbf{Approach / Metric} & MAP & MRR & MAP & MRR \\
\midrule
MSE & 0.869 & 0.911 & 0.471 & 0.513 \\
BSC & 0.801 & 0.867 & 0.495 & 0.534\\
BSC - clusters shuffle & 0.787 & 0.859 & 0.493 & 0.534 \\
BSC - random shuffle & 0.763 & 0.828 & 0.487 & 0.530\\
BSC - w/o shuffle & 0.816 & 0.884 & 0.481 & 0.532 \\
Combo BSC + MSE & \bf 0.872 & 0.912 & \bf 0.496 & \bf 0.540 \\
Triplet loss  & 0.857 & \bf 0.917 & 0.475 & 0.529  \\
\midrule
\citet{nakov-etal-2017-semeval} & \underline{0.884} & \underline{0.928} & 0.472 & 0.501\\
\bottomrule
\end{tabular}
\label{tab:sema}
\end{wraptable}

\textbf{CQA-A} The results for CQA subtask A are shown in Table~\ref{tab:sema}. A comparison with \citep{nakov-etal-2017-semeval} is not very fair, as we did not use the metadata, e.g.,~the comment position, which was crucial for the best systems. \rev{Besides, we use SBERT, which is inferior to a fine-tuned BERT.} Nevertheless, our best approach of combo training with MSE and BSC losses yielded competitive results.
We further compared different shuffling strategies. The data is ordered by questions, and keeping this order turns out to be best. That is, the model learns to distinguish positive answers for each question from manually selected negative ones and from answers to other questions. Also, note that random shuffling completely eliminates this structure, and MAP drops by 6\% absolute. Fast shuffling by 300 clusters, an advanced version of shuffling by words, improves these results. Example-based shuffling finds a data order similar to the initial one, and the quality does not degrade much.

\textbf{CQA-B} The results for CQA-B are shown in Table~\ref{tab:sema}. Again, we did not use the question position, which is a critically important feature for the best systems. We can see that the BSC loss achieved the best score, noticeably outperforming MSE and triplet losses. The experiments also demonstrate the importance of data order when training with the BSC loss. Since the dataset is small, the model overfits when the original data order is fixed.

\begin{wraptable}{R}{0.52\textwidth}
\small
\vspace{-23pt}
\centering
\caption{Results for PFCC-S.}
\begin{tabular}{l c c c}
\toprule
\textbf{Approach / Metric} & HP@1 & HP@5 & HP@50  \\
\midrule
MSE &  0.362 & 0.508 & 0.709 \\
BSC & \bf 0.673 & \bf 0.844 & 0.899 \\
BSC - 1-dim norm & 0.588 & 0.764 & 0.899 \\
BSC - no norm & 0.608 & 0.744 & 0.884 \\
BSC - random shuffle & 0.663 & 0.794 & \bf 0.915 \\
Triplet loss  & 0.668 & 0.794 & 0.899 \\
\midrule
\citet{shaar-etal-2020-known} & 0.402 & 0.653 & 0.784 \\
\bottomrule
\end{tabular}
\label{tab:pfcc-s}
\end{wraptable}

\textbf{PFCC-S} Table \ref{tab:pfcc-s} shows the results for PFCC-S (\textit{HP@k} stands for \emph{HasPositives@k}). Note that the scores from \citep{shaar-etal-2020-known} are for pre-trained SBERT without task-specific fine-tuning. We observed that even when using oversampling to improve the balance of positive examples, MSE performed worse than their results. Here, we used only positives examples to train with BSC, and normalizing by the zero dimension was the best. Overall, the approaches using BSC and triplet losses were comparable. However, the dataset size for training with the BSC loss was much smaller, which is also true for MSE. As a result, the BSC loss is faster, and preferable for this task.

\begin{wraptable}[8]{R}{0.53\textwidth}
\small
\vspace{-20pt}
\centering
\caption{Results for MRPC  and QQP.} %
\begin{tabular}{l c c}
\toprule
\textbf{Approach / Metric} & \bf MRPC (F1) & \bf QQP (F1) \\
\midrule
MSE & 89.08 & 74.29 \\
BSC  & 86.73  & 73.13 \\
Combo BSC + MSE & \bf 89.46 & \bf 75.07 \\
\midrule
\citet{thakur2020augmented} & 87.89 (88.55) & 74.97 (\underline{79.77}) \\
\bottomrule
\end{tabular}
\label{tab:mrpc}
\end{wraptable}

\textbf{MRPC} Table \ref{tab:mrpc} shows the results for MRPC. MSE outperformed the BSC loss, but combo achieved a slightly higher F1 score.

\textbf{QQP} The results for QQP are presented in Table~\ref{tab:mrpc}. We also show results for SBERT and augmented SBERT (in parentheses) from \citep{thakur2020augmented}. There score was obtained by training SBERT with MSE using another random training sample, but nonetheless, the F1 score is close to ours. The combo approach outperformed separate training with BSC or MSE.

\begin{wraptable}{R}{0.5\textwidth}
\small
\vspace{-23pt}
\centering
\caption{Results for STSb: Spearman rank correlation.}
\begin{tabular}{l c}
\toprule
\textbf{Approach / Metric} & $\rho \times 100$  \\
\midrule
MSE & 84.80 \\
BSC  & 83.26 \\
Combo BSC + MSE & 84.59 \\
Fine-tuning MSE with BSC & 84.95 \\
Fine-tuning BSC with MSE & \bf 85.71 \\
\midrule
\citet{reimers-gurevych-2019-sentence} & 84.86 \\
\bottomrule
\end{tabular}
\label{tab:sts}
\end{wraptable}

\textbf{STSb} Table \ref{tab:sts} shows the results for STS. It is the only task where combo with the BSC loss was worse than MSE. This could be due to hard negatives not appearing in the batch in any of the shuffling procedures. Moreover, we observed only marginal improvement when fine-tuning with a BSC model initially trained with MSE. However, if it was pretrained with BSC up to overfitting, fine-tuning it with MSE yielded sizable improvements.

\section{Discussion}
\label{sec:discuss}

We highlight the following observations:
\begin{itemize}
    \item Combo-training with BSC and MSE losses generally yields the best results (the only exception is STS), and it outperforms the triplet loss with advanced negative sampling.
    \item The order in which the data is presented for training can be critical, as we have seen in the cases of CQA-A and CQA-B.
    \item The use of labeled negatives examples generally improves the scores by 1-2\% absolute.
    \item Embedding normalization during training is important. Moreover, it is useful to normalize to the zero dimension (e.g., for PFCC-S).
    \item Temperature $\tau$ of order 0.1 should be used with the standard normalization, and $\tau$ of order 1-3 for coordinate normalization.
    \item An incorrect training setup may hurt the performance by more than 10\%\rev{, as was demonstrated for (\emph{i})~filtering out negative examples for which no positives were given in the dataset (Table \ref{antique}), (\emph{ii})~using poorly formed batches (highest effect in Table \ref{tab:sema}), (\emph{iii})~suboptimal normalization (Table~\ref{tab:pfcc-s}), and (\emph{iv})~wrong temperature value.}
    \item The BSC loss is more suitable for ranking tasks, but it can help for other tasks if applied as pre-training or in joint training with the MSE loss.
\end{itemize}

Selecting a loss function is important. For instance, if the model optimizes Pearson correlation, it achieves a score of 85.57 on the STS task. Thus, it outperforms \rev{almost} all considered approaches. Moreover, the combination of such a loss with BSC allows the model to achieve an F1 score of 89.88 in the MRPC task (a classification task).

Finally, we would like to draw a parallel between our work and \emph{Augmented SBERT} \citep{thakur2020augmented}. When using the BSC loss, some negatives are implicitly added to the dataset. Augmented SBERT adds new examples too and retrieves them using BM25 or Semantic Search samplings. These methods are comparable to our fast shuffling by words ($n$-grams) and to example-based shuffling, respectively. Moreover, the task-specific model is used to encode the data in both cases. However, we do not need to label such pairs with another model (cross-encoder) due to the BSC loss definition.

\section{Conclusion and Future Work}
\label{sec:concl}

We explored the idea of using a batch-softmax contrastive loss for fine-tuning large-scale pre-trained transformers to learn better task-specific sentence embeddings for pairwise sentence scoring tasks. We introduced and studied a number of variations in the calculation of the loss as well as in the overall training procedure. Our experimental results have shown sizable improvements on a number of datasets and pairwise sentence scoring tasks including ranking, classification, and regression. 

In future work, we want to explore new variations of the loss, and to gain better understanding of when to use which variation. We further plan experiments with a larger set of NLP tasks.

\bibliographystyle{iclr2022_conference}

\end{document}